\documentclass[a4paper,10pt,twoside]{article}

\usepackage{clin}        % Stylefile for CLIN Journal
\usepackage{harvard}     % Bibliography Stylefile
\usepackage{latexsym}
\usepackage{graphicx}
\usepackage[]{algorithm2e}
\usepackage{multirow}
\usepackage{url}
\usepackage{layout}
\usepackage{subfigure}
\usepackage[hidelinks]{hyperref}
\usepackage{amsmath}

\usepackage{adjustbox}
\begin{document}

\title{Selecting Parallel In-domain Sentences for Neural Machine Translation Using Monolingual Texts}

%Authors and addresses
%Example has 3 authors with 2 different affiliations
%Adapt where necessary
\author{Javad Pourmostafa Roshan Sharami$^*$ \email{j.pourmostafa@tilburguniversity.edu}\\
{\normalsize \bf Dimitar Shterionov}$^*$ \email{d.shterionov@tilburguniversity.edu}\\
{\normalsize \bf Pieter Spronck}$^*$ \email{p.spronck@tilburguniversity.edu}\\
\AND \addr{$^*$Department of Cognitive Science and Artificial Intelligence, Tilburg University, Tilburg, The Netherlands}}

\maketitle\thispagestyle{empty} % extra pagestyle command for first page

%To be filled in by the editors
%Please leave commented out
%\jmlrheading{vol}{year}{pages}{Submission date}{Publication date}{authors}
%\copyright

\begin{abstract}
Continuously-growing data volumes lead to larger generic models. Specific use-cases are usually left out, since generic models tend to perform poorly in domain-specific cases. Our work addresses this gap with a method for selecting in-domain data from generic-domain (parallel text) corpora, for the task of machine translation. The proposed method ranks sentences in parallel general-domain data according to their cosine similarity with a monolingual domain-specific data set. We then select the top $K$ sentences with the highest similarity score to train a new machine translation system tuned to the specific in-domain data. Our experimental results show that models trained on this in-domain data outperform models trained on generic or a mixture of generic and domain data. That is, our method selects high-quality domain-specific training instances at low computational cost and data size.
\end{abstract}

\section{Introduction}

A widely accepted belief among machine translation (MT) researchers and practitioners is that more training data is better. That is, the larger the training corpus is, the more robust and accurate the model can be. However, substantial amounts of parallel data are not available for all language pairs or domains of interests \cite{currey-etal-2017-copied,van-der-wees-etal-2017-dynamic,stergiadis2021multidomain}. Furthermore, data-driven machine translation systems' performance depends not only on the quantity but also on the quality of available training data \cite{fadaee-etal-2017-data}. Despite the fact that more and more training data for Machine Translation (MT) are becoming accessible every day, only those that cover the same or similar domain of interests are commonly able to boost translation quality \cite{wang-etal-2017-sentence,d84596fb74844fd7a414e2f25ad81e3e}. Hence, for domain-specific use-cases, data-driven paradigms may perform poorly when trained on general-domain data, regardless of the size of the corpus. Training MT systems on large amounts of data in many cases uses substantial amounts of resources such as memory and time which is the undesired effect of striving to boost the performance of MT systems. As such, it is paramount to be able to train systems on high-quality domain-specific data. We, therefore, face a two-sided challenge: (i) what is high-quality, in-domain data and (ii) what amounts of parallel, in-domain data are necessary to achieve state-of-the-art MT quality at low computational and data capacities.

To address these challenges, the research community has made many efforts to improve MT performance through Domain Adaptation (DA) techniques.
DA for MT has versatile definitions, but we primarily follow \citeasnoun{chu-wang-2018-survey}, who state that \emph{DA would be employing out-of-domain parallel corpora and in-domain monolingual corpora to improve in-domain translation}. Among other definitions, \citeasnoun{saunders2021domain} defined DA as any scheme that aims to level up translation's performance from an existing system for a specific topic or genre language. Studies in this area are mainly divided into two categories, namely (i) data-centric and (ii) model-centric~\cite{chu-wang-2018-survey}. The data-centric category includes methods that operate at the corpus / data level by selecting, generating, joining, or weighting sentences or data sets for training purposes. This category selects or generates the domain-related sentences from the general domain using existing in-domain / domain-specific data. In contrast, in the model-centric category, studies mostly fall into the areas aiming to alter the usual function of models. This is usually fulfilled through mixing, fine-tuning and reordering models or weighting objective functions, or applying regularization techniques.

Our proposed methodology falls into a data-centric category and is specifically considered a data selection method. However, a few previous studies have investigated the generation of parallel in-domain sentences for MT. Their limitations motivate us to expand this area of research by proposing a novel data selection algorithm for collecting in-domain training data. %and each has its limitations which results in motivation for us to expand this area of research. i.e., selection of in-domain training data employing a data selection approach. 
In this regard, we aim to improve in-domain translation in low-resource scenarios by selecting in-domain sentences from out-of-domain corpora, then possibly employing Domain Adaptation (DA) for Neural Machine Translation (NMT) leveraging both out-of-domain parallel corpora and in-domain monolingual data. In essence, our proposed approach leads to one main contribution: \emph{a language-agnostic data selection method for generating a parallel in-domain corpus using monolingual (untranslated) domain-specific corpora}. Monolingual corpora are often abundant and can easily be collected without requiring any translations or further sentence alignments. This has two consequences. The first one is related to the proportion of high-quality data to the number of in-domain sentences. In particular, our method generates fewer but of higher quality sentences with the same or at least competitive performance on NMT systems. The second one is the reduction of training time. This is a consequence of the fact that less data is used for training an NMT system.

\begin{figure}[ht]
	\centering 
	\includegraphics[width=3.2in]{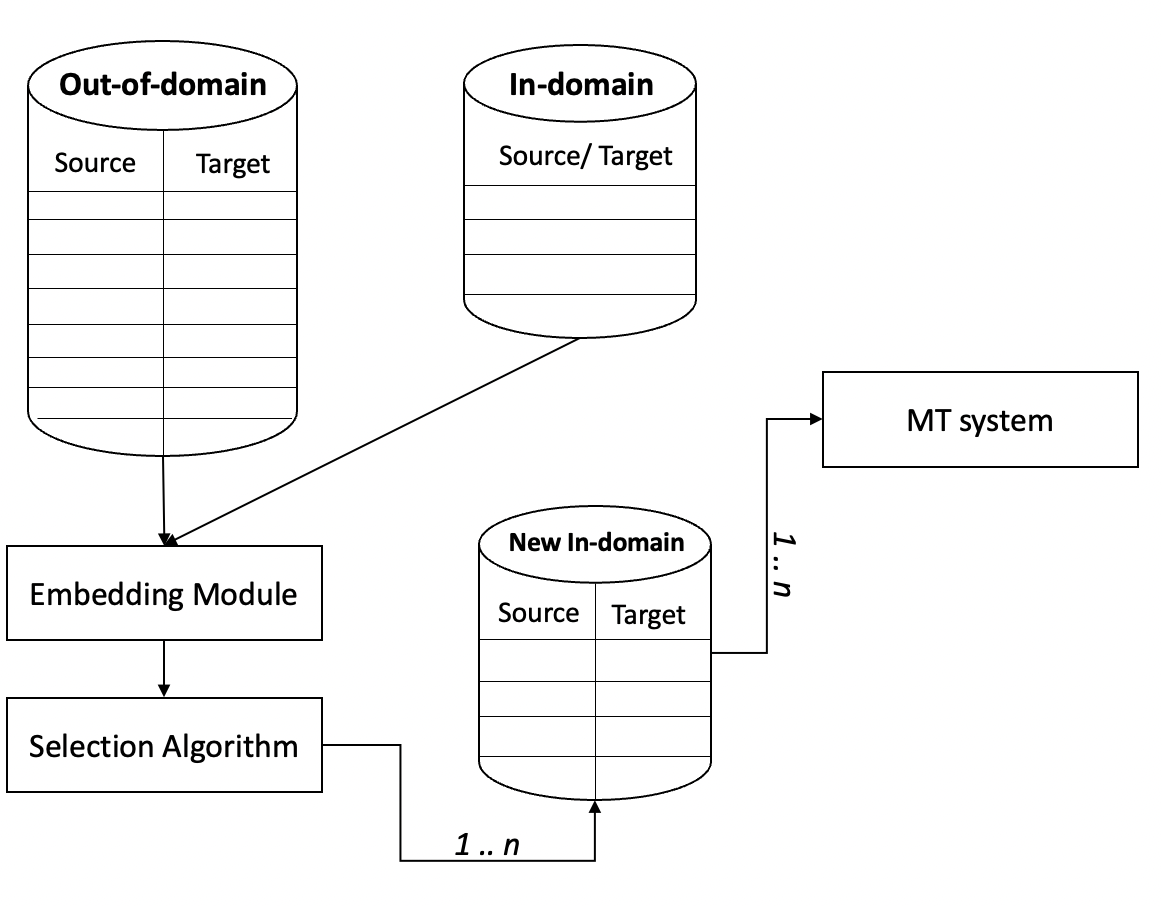} 
	\caption{An overview of the proposed methodology. $1 .. n$ indicates that the algorithm can select between $1$ and $n$ sentences, where $n$ is an arbitrary number.} 
	\label{fig:method_overview}
\end{figure}

In this paper, we created a large parallel out-of-domain corpus (e.g. EN$\Rightarrow$FR), using several smaller corpora. This is mainly because a general-domain corpus should be sufficiently broad in terms of sentence diversity such that it increases the number of in-domain data. Likewise, a monolingual in-domain corpus containing in-domain sentences (either EN or FR) was utilized. Then, both selected corpora were embedded to be used for further analysis. A dimensionality reduction technique, i.e. Principal Component Analysis (PCA), was applied to them to mitigate the computational costs. Dimensionality reduction will be discussed in detail in Section~\ref{sec:dimensionality_reduction}. The in-domain vectors were compared to out-of-domain sentences, then similar embedded vectors were ranked in descending order to generate an in-domain parallel corpus. The ranked sentences were also mixed to increase the amount of training data. Eventually, each one was fed into the MT systems to be trained and subsequently, the best translation indicates the best quality of an in-domain parallel corpus. This abstract perception is depicted in Figure \ref{fig:method_overview}.

The paper is organized as follows. We first cover the related work and define specialized terminology in Section~\ref{sec:related_work}. In Section~\ref{sec:data_selection}, our proposed strategy regarding data selection is presented. Next, empirical evaluation including details about the train and test data sets, systems specifications, baselines and results are shown in Section~\ref{sec:experiments}. Section~\ref{sec:discussion} gives further insights into the proposed method. Section~\ref{sec:conclusions} concludes the paper.

\section{Related Work}
\label{sec:related_work}
There is a significant volume of research in DA for MT paradigms. However, to the best of our knowledge, few prior studies have been conducted particularly on selecting in-domain sentences efficiently and then exploiting them to improve the in-domain translation. That is, for the related work we selected data-centric and more specifically data selection papers that were closely related to our research. \citeasnoun{Luong2015StanfordNM} did major work in this area when they adapted an existing English-German deep Long Short-Term Memory (LSTM) model by training it for additional 12 epochs on new domain data in the same languages; the original training data is general-domain, while the one used for adaptation is from the conversational domain. This DA approach led to an increase of 3.8 BLEU points compared to the original model (25.6 to 29.4) without further training. Similarly, \citeasnoun{zoph-etal-2016-transfer} proposed a transfer learning method for low-resource language pairs.

Among previous works, the research presented in \cite{wang-etal-2017-sentence} is very similar to our work in terms of the intuition behind the selection methodology. Their method selects in-domain data based on similarity scores computed over embeddings drawn from an NMT system trained on in and out of domain data. This approach has several limitations related to the fact that it relies on a particular NMT system that needs to be trained on both in- and out-of-domain data. That is, the complexity of their approach makes it rather difficult to employ in practice. Furthermore, as it relies on the embeddings of a particular NMT, it implies that a (language-specific) NMT system needs to be available or trained which may add computational or economic overhead.

\citeasnoun{axelrod-etal-2011-domain} proposed a DA approach using data selection which is a common baseline for many contemporary works. This is mainly because their work for the first time introduced the concept of domain adaptation in MT. They selected and ranked sentences with three cross-entropy based methods for the task of SMT. They also showed that all three methods presented in their paper outperformed the general-domain model. In the same direction, \citeasnoun{Chen2016BilingualMF} presented another new data selection technique employing Semi-Supervised Convolutional Neural Networks based on bitokens (Bi-SSCNN). The method they proposed only requires a small amount of in-domain data to train the selection model. Suggested methods were tested on two translation tasks (Chinese-to-English and Arabic-to-English) and showed that the Bi-SSCNN is more functional than other approaches in terms of averting noisy sentence pairs. We compare our approach to the aforementioned models (among others). In Section~\ref{sec:experiments} we outline those models and present further details and comparisons. 

With respect to data selection studies, \citeasnoun{van-der-wees-etal-2017-dynamic} also investigated a method, called dynamic data selection, to discern whether it is feasible to improve NMT performance. Their method sifts through all training data between training epochs subsequently and reduced the training data size by selecting sentence pairs most relevant to the translation task. By doing so, unlike fixed training data, the training becomes a gradual fine-tuning process, which iterates over different training subsets made. \citeasnoun{chu-etal-2017-empirical} proposed a novel DA method, called mixed fine-tuning, incorporating fine-tuning into multi-domain \cite{sennrich-etal-2016-controlling,Kobus_2017} for NMT. In the context of the corpora they experimented with, their fine-tuning method on a mix of in-domain and out-of-domain solves the problem of overfitting. 

It is also worth mentioning the work of \citeasnoun{aharoni-goldberg-2020-unsupervised} which proposes two methods for data selection based on unsupervised language models. The first one computes the centroid of the in-domain data and selects the samples that are nearest to the centroid, according to their cosine similarity. The second method is based on a binary classifier trained on in-domain sentences and a random negative sample (i.e., general-domain) of sentences. They evaluate the proposed approach on 5 domains and compare it to the work of \citeasnoun{moore-lewis-2010-intelligent}.

\section{Data Selection Method}
\label{sec:data_selection}
%We present a method to select and rank in-domain subsets of an out-of-domain corpus, in an attempt to boost in-domain translation performance. 
Our method ranks sentences in a general-domain (or out-of-domain) data set according to their similarity with an in-domain data set. This in-domain data set is monolingual; if a parallel corpus is provided only the source or target side is to be used. Once the sentences are ranked, we can then extract the top K sentences with the highest score, i.e. ranked the highest and use those for training a new MT system. According to our architecture, initially the input data, both in-domain and out-of-domain is converted into embedding vectors before. Our method then computes the similarity between these vectors and uses the similarity score for ranking and consecutively, for selection. The embedding space we use, Sentence BERT~\cite{reimers-gurevych-2019-sentence}, is of high dimensionality. As such, these vectors become quite large to be processed effectively. Our architecture exploits PCA to minimize them optimally which will allow the next step -- semantic search and ranking -- to be conducted as efficiently as possible.
 
\subsection{Sentence Embedding and Dimensionality Reduction}
\label{sec:dimensionality_reduction}
Word representation is a rich resource for gaining information for downstream tasks such as classification, entailment, translation, etc. Natural Language Processing (NLP), including MT, models greatly depend on the representation of input \cite{10.1145/3434237}. While the myriad of categorical and fixed methods such as Bag-of-Words (BOW), Continuous BOW model, Skip-Gram Model \cite{mikolov2013efficient}, FastText \cite{bojanowski2017enriching} have been employed in this regard, nowadays researchers tend to benefit more from unsupervised contextual word representations architectures and in particular transformer-based language models \cite{vaswani2017attention}. The main reasons are that (i) these models keep the full context of the input and (ii) they reduce the computational time for most NLP tasks. These directly align with our research objectives: to be able to select semantically similar in-domain sentences and to reduce MT training time as well as similarity computational time. Hence, efficient sentence embedding plays a major role in our work.

There exist several transformer-based language models, such as, BERT \cite{devlin-etal-2019-bert} and RoBERTa \cite{liu2019roberta} that have set a state-of-the-art baseline \cite{cer-etal-2017-semeval}. However, for tasks like Semantic Textual Similarity (STS), Sentence-BERT (SBERT) \cite{reimers-gurevych-2019-sentence} recently showed a better performance. It is a modification of the pre-trained BERT network that employs Siamese \cite{10.5555/2987189.2987282} and triplet network structures \cite{Schroff_2015}, capable of capturing meaning-related relationships which allow assessing the degree to which two sentences are semantically similar at reduced computational costs. We note that SBERT is only used in our research for embedding words and not as input for MT.

By default, the SBERT base model embeddings output has 768 dimensions. Sentences encoded using 768-dimensional vectors require a substantial amount of memory to be stored. For example, an out-of-domain corpus containing 31 million sentences generates a $31M\times768$ embedding matrix which is computationally expensive. The size of these large vectors does not only require substantial physical memory but also considerably increases the time for computing the semantic similarity (i.e. the STS task), where we need to cautiously mitigate the cost of computing semantic search between in-domain and out-of-domain sentences. Furthermore, considering that embedding vectors are used for semantic search, they should load to GPU memory to speed up the search process. This leads to a dearth of memory if our embedding output size is large (which is true in our case).

In order to mitigate these issues, we decided to work with smaller sized vectors. However, reducing the vector dimensions must be done in a conscious way, in order not to lose important information. To do so, we employ PCA~\cite{Jolliffe2011} as a pooling method into the last embedding neural layers. We select the 32 principal components as our output features. That is, PCA acts as the final layer of our selection network which allows us to reduce the 768-dimensional vectors to 32-dimensional ones. In general, PCA components are equivalent to the output features and are easy to append to any SBERT pre-trained model. 

To employ the PCA idea in SBERT, we only selected and shuffled 500 K sentences of experimental data sets to train a PCA model.
A pre-trained model called 'stsb-xlm-r-multilingual'\footnote{\url{https://huggingface.co/sentence-transformers/stsb-xlm-r-multilingual}} (trained on over 50 languages), has been selected and subsequently 32 components appended to that. Finally, the reduced model is saved to use for word embedding. 

\subsection{Semantic Search and Ranking In-domain Data}
\label{sec:semanticssearch}
Once the input data was reduced and embedded, we employed semantic search to identify general-domain sentences that are similar to in-domain data. This idea is mainly inspired by \cite{moore-lewis-2010-intelligent,axelrod-etal-2011-domain,duh-etal-2013-adaptation,wang-etal-2017-sentence}. In particular, we assume in-domain and out-of-domain sentences as search queries and document entries respectively. In our assumption, queries are responsible to find the most relevant embedding vectors from entries. The proximity search can be fulfilled by various distance similarity measures methods, such as cosine similarity, Manhattan distance, Euclidean distance, etc. In our research, we intuitively used cosine similarity. However, we could not directly apply that on our data sets using CPU as such an exhaustive search (each in-domain sentence compared to each out-of-domain sentence) was expensive, specifically in terms of computational time. To resolve this issue, we used a GPU implementation of cosine similarity in the PyTorch library. Once the cosine similarity scores were computed over embedding vectors, the torch.topk\footnote{\url{https://pytorch.org/docs/stable/generated/torch.topk.html}} is called. By calling this function the n largest / most similar elements, as well as indices of the given in-domain or out-of-domain embedded tensors, are returned.

Let $S$, $E$, $q$ and $d$ denote in-domain corpus, out-of-domain corpus, a vectorized search query and an embedded document entry, respectively, where $q\in S$ and $d\in E$. Let $k$ and $l$ be the number of sentences in the given corpora. Thus $S=\left \{q_1,q_2,\text{...},q_k\right \}$, $E=\left \{d_1,d_2,\text{...},d_l \right \}$ and $k\ll l$. According to these definitions, the cosine similarity is defined by Equation \ref{eq1} in our data selection method, where 32 is the number of dimensions. 

\begin{equation}
\centering
\label{eq1}
cos(\vec{q}, \vec{d}) = \frac{\vec{q}\cdot \vec{d}}{|\vec{q}||\vec{d}|} = \frac{\vec{q}}{|\vec{q}|}\cdot \frac{\vec{d}}{|\vec{d}|} = \frac{\sum_{i=1}^{32}q_{i}d_{i}}{\sqrt{\sum_{i=1}^{32}{q_{i}}^{2}}\sqrt{\sum_{i=1}^{32}{d_{i}}^{2}}}
\end{equation}

Based on the similarity measurement defined in Equation~\ref{eq1}, we rank our sentences and pick the top $n$ ($n = 6$ in our experiments) out-of-domain sentences, which are sorted in descending order, to build pseudo in-domain sub-corpora. Considering that sentences are chosen from an out-of-domain corpus and are distinct bitexts, no further operation is required before feeding those into an NMT system.

In summary, our data selection method has four major steps:
\begin{itemize}
    \item Step 1, the input data is converted into vectors using the embedding unit.
    \item Step 2, the vectors' dimensions are reduced to the lower dimensions.
    \item Step 3, we compute the similarity scores between each in-domain and out-of-domain vector.
    \item Step 4, vectors pairs are ranked according to their similarity score achieved from step 3.
\end{itemize}

\section{Experiments}
\label{sec:experiments}

To test our data selection performance in generating parallel in-domain data, we conducted experiments with English-French data and compared our results to different systems' results. These systems are divided mainly into two categories: (i) for the first category, we trained some models for specific purposes, such as having an in-domain NMT model that only uses the original / given in-domain data for training. i.e., systems that are not trained with the selected data generated by our data selection method. This shows how well our data selection algorithm worked in terms of helping the model to reach the maximum possible translation performance. It is noteworthy that the systems in the first category are not considered baselines. 
(ii) For the second category, we chose previous researchers' MT systems that were trained using their selected data. These systems after being trained on the selected in-domain data are usually re-trained on the original / given in-domain data to increase the translation performance. Although most systems in this category used DA / re-training in their work, we considered them as actual baselines. This is, first, because there is not much previous work that only uses in-domain parallel data to train MT systems. Second, we aim to evaluate the quality of our generated data to see if it helps the models to improve the translation quality without re-training, i.e., as a stand-alone corpus.

\subsection{Data}
\label{sec:data}
\paragraph{In- and out-of-domain data sets.}
The data we experimented with is called IWSLT\footnote{International Workshop on Spoken Language Translation} 2014 corpus \cite{Cettolo2015ReportOT}, collected through the TED talks. To validate our models during training time and test the models' performance, we used one development set (dev2010) and two test sets (test2010 and test2011), respectively. In addition to an in-domain corpus, we combined a collection of WMT corpora\footnote{\url{http://statmt.org/wmt15/translation-task.html}} including Common Crawl, Europarlv7, News Commentary v10 and United Nations (UN) \cite{Koehn2005EuroparlAP,tiedemann-2012-parallel} to create a large out-of-domain corpus. IWSLT 2014 and WMT are commonly used in the context of DA as an in-domain data set \cite{axelrod-etal-2011-domain,Luong2015StanfordNM,Chen2016BilingualMF,wang-etal-2017-sentence}, which allows for better replicability. Data statistics are shown in Table \ref{tab:corpora}.

\begin{table}[ht]
\renewcommand{\arraystretch}{1.40}
\centering
\begin{adjustbox}{width=290pt,center}
\begin{tabular}{c|c|c|c}
\hline
\hline
\textbf{EN-FR}           & \textbf{Name}                & \textbf{Sentences} & \textbf{Talks} \\ \hline
TED training (in-domain) & \multirow{4}{*}{IWSLT 2014}  & 179K                  & 1415           \\ \cline{1-1} \cline{3-4} 
TED dev2010              &                              & 887                   & 8              \\ \cline{1-1} \cline{3-4} 
TED test2010             &                              & 1664                  & 11             \\ \cline{1-1} \cline{3-4} 
TED test2011             &                              & 818                   & 8              \\ \hline
\multirow{4}{*}{WMT training (out-of-domain)} & Common Crawl & 3.25M & \multirow{4}{*}{$\sim$31M} \\ \cline{2-3}
                         & Europarl v7                  & 2M                 &                \\ \cline{2-3}
                         & News Commentary v 10         & 200K               &                \\ \cline{2-3}
                         & United Nations (UN)          & 25.8M              &                \\ \hline \hline
\end{tabular}
\end{adjustbox}
\caption{Summary of in-domain and out-of-domain data sets}\label{tab:corpora}
\end{table}

\paragraph{Selected data.} We used our method (see Section~\ref{sec:semanticssearch}) and the monolingual in-domain data set, i.e. TED, to extract the most similar subsets of in-domain data from out-of-domain data, i.e, WMT. Since the data we extract is not specifically compiled as (authentic) in-domain data, but rather automatically generated based on similarities, it is referred to as \emph{pseudo} in-domain data~\cite{zhang-xiong-2018-sentence}. In this paper, to distinguish between authentic and pseudo in-domain data, we refer to the latter as selected in-domain data as this reflects the origin of the data. 

\begin{figure}[ht]
	\centering 
	\includegraphics[width=4.2in]{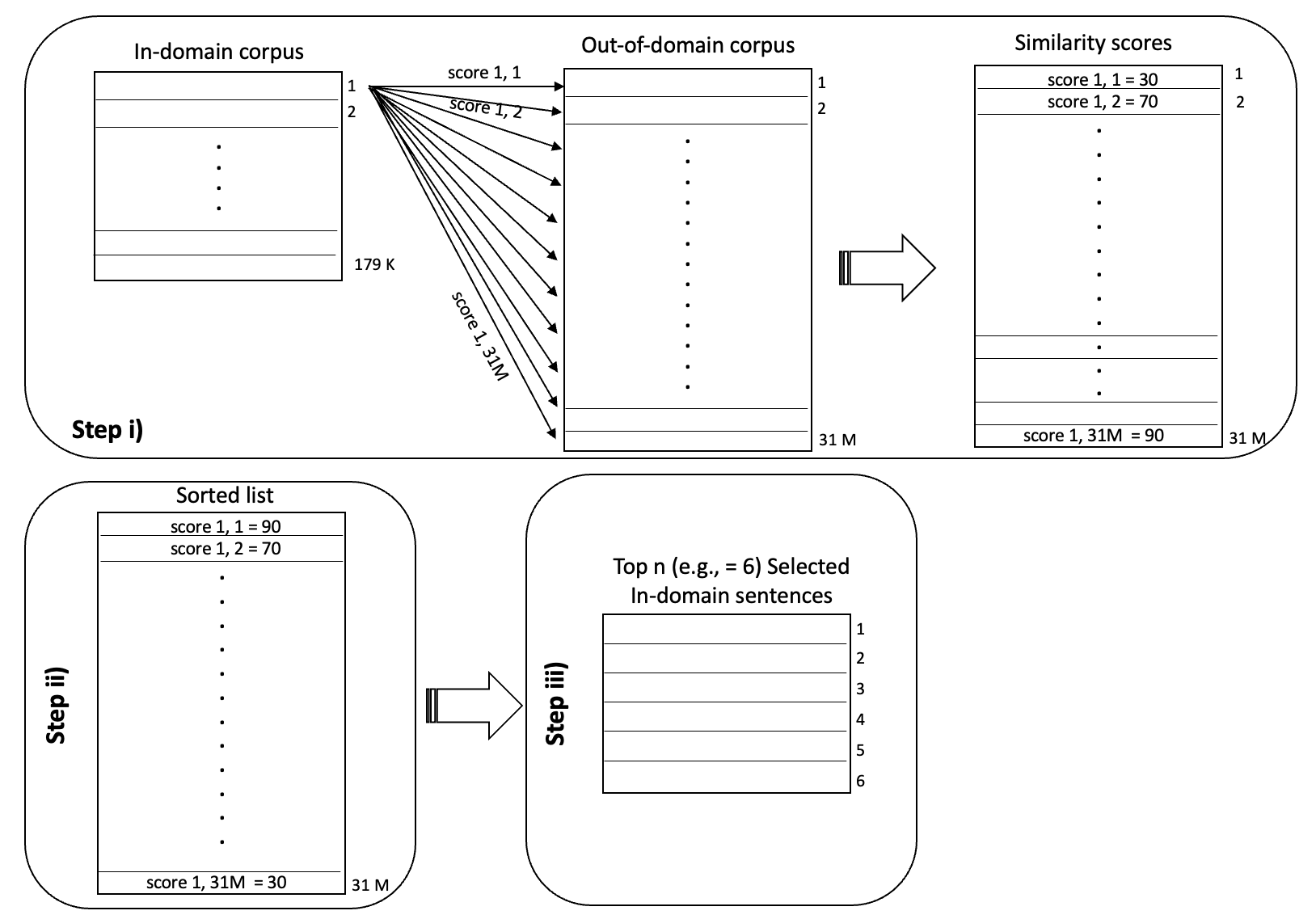} 
	\caption{An iteration of selecting in-domain data.} 
	\label{fig:dataselection}
\end{figure}

In particular, we created six (sub-)corpora based on the sentence ranks determined by the data selection method. To create them, (i) we choose one in-domain sentence and compute its similarity score with every single out-of-domain data point, i.e. one-to-many. So, for each in-domain sentence, we obtain a score list with the size of out-of-domain data, i.e. 31M. (ii) Afterward, the out-of-domain sentences are descendingly sorted according to the similarity scores achieved in step (i); (iii) We only select first n ($n=6$) out-of-domain sentences from the list generated in step (ii). We repeat all aforementioned steps for every single in-domain entry, i.e., 179K. This procedure outputs a $179k\times 6$ matrix according to our input data. Figure \ref{fig:dataselection} shows one iteration of data selection.

Table \ref{tab:DSoutput} shows an example of possible outputs for the proposed data selection algorithm given a monolingual in-domain query, where generated sentences were sorted from the highest score (top1) to the lowest one (top6).

\begin{table*}[ht]
\centering
\begin{adjustbox}{width=\textwidth,center}
\renewcommand{\arraystretch}{1.7}
\begin{tabular}{rl|c|l|}
\cline{3-4}
 &
   &
  \multicolumn{2}{l|}{Score (/100)} \\ \hline
\multicolumn{1}{|r}{Monolingual in-domain ($q_i$):} &
  It can be a very complicated thing, the ocean. &
  \multicolumn{2}{c|}{-} \\ \hline
\multicolumn{1}{|c}{Top1–parallel in-domain ($top_{i0}$):} &
  \begin{tabular}[c]{@{}l@{}}EN: Ocean affairs are sensitive and complex.\\ FR: Les affaires maritimes sont délicates et complexes.\end{tabular} &
  \multicolumn{2}{c|}{90.10} \\ \hline
\multicolumn{1}{|r}{Top2–parallel in-domain ($top_{i1}$):} &
  \begin{tabular}[c]{@{}l@{}}EN: This is a dangerous position to be in if the sea is running high.\\ FR: Ainsi, le capitaine peut prendre les effets du capitaine du navire pris, le chirurgien ...\end{tabular} &
  \multicolumn{2}{c|}{86.80} \\ \hline
\multicolumn{1}{|r}{Top3–parallel in-domain ($top_{i2}$):} &
  \begin{tabular}[c]{@{}l@{}}EN: Rip currents and undertow are common, dangerous conditions along ocean beaches.\\ FR: Déchirez les courants et les baïnes sont des conditions communes et dangereuses le long des plages d'océan.\end{tabular} &
  \multicolumn{2}{c|}{86.60} \\ \hline
\multicolumn{1}{|r}{Top4–parallel in-domain ($top_{i3}$):} &
  \begin{tabular}[c]{@{}l@{}}EN: Moving with the waves can be dangerous.\\ FR: Il est dangereux de progresser avec la vague.\end{tabular} &
  \multicolumn{2}{c|}{86.13} \\ \hline
\multicolumn{1}{|r}{Top5–parallel in-domain ($top_{i4}$):} &
  \begin{tabular}[c]{@{}l@{}}EN: Obstacles in the water are particularly dangerous when coupled with currents.\\ FR: Les obstacles dans l’eau sont avant tout dangereux par rapport au courant.\end{tabular} &
  \multicolumn{2}{c|}{85.96} \\ \hline
\multicolumn{1}{|r}{Top6–parallel in-domain ($top_{i5}$):} &
  \begin{tabular}[c]{@{}l@{}}EN: This problem affects not only small islands, but also large islands and countries with extensive coastlines.\\ FR: Ce problème concerne non seulement les petites îles, mais aussi les grandes îles ....\end{tabular} &
  \multicolumn{2}{c|}{85.76} \\ \hline
\end{tabular}
\end{adjustbox}
\caption{An example of data selection output}
\label{tab:DSoutput}
\end{table*}

\begin{figure}[ht]
	\centering 
	\includegraphics[width=3.5in]{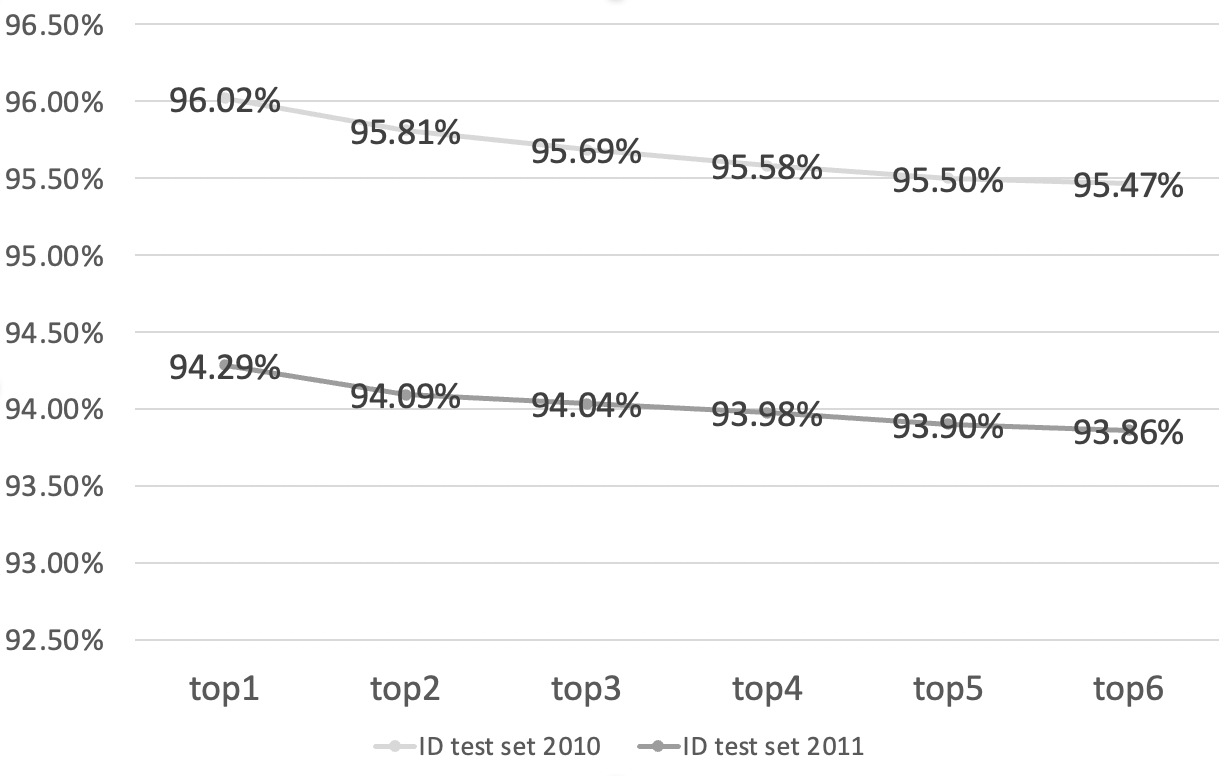} 
	\caption{The difference of selected sub-corpora to in-domain test sets} 
	\label{fig:closeness}
\end{figure}

To show the effectiveness of our semantic search and ranking idea, we found the centroids of selected sub-corpora, then compared them to the in-domain test sets' centroids. The centroid is a multidimensional vector, calculated as the average of all other vectors. That is, it is a vector around which other vectors are distributed. Figure~\ref{fig:closeness} depicts a gradual drop of similarity score from top1 to top6 for both test sets. To increase the performance of in-domain translation in terms of diversity richness, each sub-corpus is combined with all preceding bitext data (like a stack). For instance, top6 comprises top5, top4, top3, top2, top1; top5 holds top4, top3, top2, top1 and so forth. In that way, that last corpus encompasses all in-domain sentences.

\subsection{NMT System Description}
We used the OpenNMT-py\footnote{\url{https://opennmt.net/OpenNMT-py/}} framework \cite{klein-etal-2017-opennmt} for training our NMT models. We trained transformer models \cite{vaswani2017attention} for a maximum of 200K steps; intermediate models were saved and validated every 1000 steps until reached convergence unless being stuck in the early stopping condition (10 validation steps with no improvements). The parameter setup stated in Table~\ref{tbl:hyperparameters} was used. We performed a large set of preliminary experiments to determine these values. These experiments span beyond the scope of this paper and as such are not further discussed, except for the batch size. The batch size and the learning rate have a direct connection to how the network learns, especially in different conditions of data sparsity / availability. As such, we conducted a thorough investigation of both and we found that using batch size 512 and the default learning rate in the OpenNMT was particularly impactful. Our results are summarised in Section~\ref{sec:batch_size}.

\begin{table}[h]
    \begin{minipage}[t]{0.42\textwidth}
        \centering
        {\small \setlength\tabcolsep{5pt} 
        \begin{tabular}{l|c}\hline\hline
            word embedding dimension & 512\\\hline
            number of transformer layers & 6\\\hline
            transformer-ff size & 2048\\\hline
            number of heads & 8\\\hline
            batch size & 512\\\hline
            batch type & tokens\\\hline
        \end{tabular}}
    \end{minipage}
    \begin{minipage}[t]{0.48\linewidth}
        \centering
        {\small \setlength\tabcolsep{5pt} 
        \begin{tabular}{l|c}\hline
            maximum sequence length & 150\\\hline
            learning optimizer & Adam~\cite{kingma2014adam}\\\hline
            optimizer learning rate & 2\\\hline
            optimizer beta1 and beta2 & 0.9 and 0.998\\\hline
            beam size for translation & 6\\\hline\hline
            \multicolumn{2}{c}{}\\
        \end{tabular}}
    \end{minipage}
    \caption{Hyperparameters used for training our NMT models.}
    \label{tbl:hyperparameters}
\end{table}

To run all NMT systems effectively, we also set other hyperparameters, as suggested by the OpenNMT-py community to simulate Google's default setup \cite{vaswani2017attention}. The NMT training was distributed over three NVIDIA Tesla V100 GPUs. We encoded all data, including rare and unknown words as a sequence of subword units using Byte Pair Encoding (BPE) \cite{sennrich-etal-2016-neural}. We built our vocabularies for both source and target languages separately. By doing so, our systems are not only capable of translating out-of-vocabulary tokens, but also rare in-vocabulary ones. The number of merge operations for BPE is 50,000 for all sub-corpora, i.e., For each selected sub-corpora, a separate BPE model was created. Vocabulary sizes of selected in-domain data are shown in Table \ref{tbl:vocab}.

\begin{table}[ht]
\centering
\begin{adjustbox}{width=280pt,center}
\renewcommand{\arraystretch}{1.3}
\begin{tabular}{l|c|c}
\hline
\hline
\multicolumn{1}{c|}{\multirow{2}{*}{\textbf{Selected In-domain Data sets}}} & \multicolumn{2}{c}{\textbf{Vocabulary with BPE}} \\ \cline{2-3} 
\multicolumn{1}{c|}{} & \textbf{Source (EN)} & \textbf{Target (FR)} \\ \hline
Top1 & 48,956 & 49,281 \\ \hline
Top2 + top1 & 49,896 & 50,055 \\ \hline
Top3 + top2 + ... & 50,299 & 50,391 \\ \hline
Top4 + top3 + ... & 50,596 & 50,568 \\ \hline
Top5 + top4 + ... & 50,759 & 50,720 \\ \hline
Top6 + top5 + ... & 50,874 & 50,894 \\ \hline \hline
\end{tabular}
\end{adjustbox}
\caption{Vocabulary sizes after applying BPE}
\label{tbl:vocab}
\end{table}

\subsection{Compared Systems}
To show the effectiveness of the proposed data selection method in terms of the quality of generating parallel in-domain data, as mentioned earlier, we compared our results with two disparate categories of NMT systems as follows. First, we used the available in-domain and out-of-domain data as well as their mixture to establish the following systems: (i) S1:ID -- NMT trained on in-domain only; (ii) S2:OOD -- NMT trained on out-of-domain data; and (iii) S3:ID+OOD -- NMT trained on the combination of in- and out-of-domain data. It is noteworthy that here, we employed a bitext in-domain data to demonstrate our selection method's productivity, however, the proposed method uses a monolingual in-domain corpus in a real use case.

System S1, S2 and S3 are not intended as baselines, but points of comparisons for different purposes: (i) system S1 resembles the best possible translation quality according to the given in-domain data; (ii) system S2 is trained on a large corpus without including domain-relevant data, thus it is a generic-domain MT system; and (iii) system S3 is trained on a mixture of a large generic-domain corpus (the same as for S2) and domain-relevant data and aims to test the impact of the in-domain data on the translation performance, resembling a domain-adapted model. 
Second, we compared our NMT systems (based on the proposed data selection method) to four reliable previous NMT DA methods, which we refer to as baselines. These are B4:Luong \cite{Luong2015StanfordNM}, B5:Axelrod \cite{axelrod-etal-2011-domain}, B6:Chen \cite{Chen2016BilingualMF}, B7:Wang\cite{wang-etal-2017-sentence} as mentioned in Section \ref{sec:related_work}. While \citeasnoun{wang-etal-2017-sentence} proposed three different models, we only refer to the best one.

\subsection{Results and Analysis}
\label{sec:results_analayses}

The performance of our MT systems is reported on two test sets with case insensitive BLEU \cite{10.3115/1073083.1073135}, TER \cite{snover-etal-2006-study} and chrF2 \cite{popovic-2015-chrf} metrics, which are implemented by sacreBLEU \cite{post-2018-call}. We also analyzed the results for any statistically significant differences (see Section \ref{subsec:ss}). The vocabulary was built by employing the IWSLT dev set and each selected in-domain corpus. Note that according to Table \ref{tab:indomain_data_selection} data, S1 achieved the highest BLEU score between compared systems in both test sets. We considered it as a saturation point. That is, we aim to generate parallel in-domain sentences as good as S1.

\begin{table}[ht]
\renewcommand{\arraystretch}{1.3}
\centering
\begin{adjustbox}{width=380pt,center}
\begin{tabular}{l|l|c|c|c|c|c|c}
\hline
\hline
\multirow{2}{*}{Systems} &
  \multirow{2}{*}{\begin{tabular}[c]{@{}l@{}}Number of \\ Sentences\end{tabular}} &
  \multicolumn{3}{c|}{NMT- Test Set 2010} &
  \multicolumn{3}{c}{NMT- Test Set 2011} \\ \cline{3-8} 
 &
   &
  \multicolumn{1}{l|}{BLEU$\uparrow$} &
  \multicolumn{1}{l|}{TER$\downarrow$} &
  \multicolumn{1}{l|}{CHRF2$\uparrow$}  &
  \multicolumn{1}{l|}{BLEU$\uparrow$} &
  \multicolumn{1}{l|}{TER$\downarrow$} &
  \multicolumn{1}{l}{CHRF2$\uparrow$} \\ \hline
S1:ID         & 179K      & 31.9           & 56.6           & 57.0        & 38.3          & 49.7        & 61.0          \\
S2:OOD        & 31.0M     & 25.8           & 66.1           & 53.0        & 30.7          & 59.3        & 47.0          \\
S3:ID+OOD     & 31.1M     & 26.0           & 62.9           & 54.0        & 30.9          & 56.8        & 58.0          \\
\hline
B4:Luong      & 17.9M     & 32.2           & N/A            & N/A         & 35.0          & N/A         & N/A           \\
B5:Axelrod    & 9.0M      & 32.2           & N/A            & N/A         & 35.5          & N/A         & N/A           \\
B6:Chen       & 7.3M      & 30.3           & N/A            & N/A         & 33.8          & N/A         & N/A           \\
B7:Wang       & 3.7-7.3M  & 32.8           & N/A            & N/A         & 36.5          & N/A         & N/A           \\
\hline
Top1          & 179K      & 21.8           & 69.8           & 50.0          & 25.6          & 64.0          & 53.0            \\
Top2+top1+... & 358K      & 26.7           & 63.4           & 54.0          & 31.3          & 57.1        & 57.0            \\
Top3+top2+... & 537K      & 29.1           & 60.4           & 56.0          & 34.3          & 53.9        & 60.0            \\
Top4+top3+... & 716K      & 30.7           & 59.5           & 57.0          & 35.6          & 52.6        & 61.0            \\
Top5+top4+... & 895K      & 30.9           & 59.1           & 57.0          & \textbf{36.7} & \textbf{51.5} & \textbf{62.0}  \\
Top6+top5+... & 1.0M      & \textbf{31.3}  & \textbf{58.3}  & \textbf{58.0} & 36.5          & 50.9        & 62.0            \\
\hline \hline
\end{tabular}
\end{adjustbox}
\caption{Evaluation scores for NMT system. These include NMT systems trained on in-domain (S1, B4, B5, B6, B7 and Top1 .. Top6+top5+...), on out-of-domain data (S2) and on mixture of both (S3).}
\label{tab:indomain_data_selection}
%\vspace{-3mm}%Put here to reduce too much white space after your table
\end{table}

Even though system S2 employed an enormous corpus with almost 31M sentences, it could not perform well for in-domain translation. System S3 had a minor improvement (0.2 BLEU points) after mixing with ID, yet less than S1 performance. This shows that more training data is not always sufficient for in-domain translation and even might lead to an under-fitted model. Furthermore, model S3 was biased because of too much out-of-domain data.

The proposed sub-corpora (top1 to top6) were employed to train the NMT models. In this regard, we started from top1, then continued with the mixture of other sub-corpora. In the beginning, there was a sharp improvement from top1 to top2 (+4.9 BLEU points for test set 2010 and +5.7 BLEU points for test set 2011). This growth trend continued until its saturation point in top5 and top6, where the performance started degrading. Given this point, we only selected six sub-corpora to achieve the maximum translation performance. Hence, top5 and top6 obtained the highest BLEU scores among the selected sub-corpora, 36.7 for test set 2011 and 31.3 for test set 2010, respectively.

Alongside BLEU, we also evaluated our models with TER and chrF2. According to TER, system top1 achieved the highest score among all systems trained on mixed sub-corpora. This would imply that this system would require the most post-editing effort. The TER score for top2 mixed with top1 dropped by 6.4 and 6.9 for test sets 2010 and 2011, respectively. This improvement (decrease) per each mixing operation continued until the system top6+top5+... but with a smaller amount of drop. According to the chrF2 metric, there was an increasing trend from top1 to top6+top5+... for test set 2010, where the last sub-corpus obtained 58.0 scores, while for test set 2011, two last systems (top5+top4+... and top6+top5+...) achieved the same chrF2 score (62.0). 

The three metrics have consistently improving trends, which supports the observation that, while more data implies an increase in quality, there is a point after which no (significant) improvements are possible. 

Although systems B4, B5, B6 and B7 were used inherently for domain adaptation of NMT systems (fine-tuned on a large corpus), top5 outperformed the best of them (B7) for test set 2011 without retraining on such enormous generic corpus. That is, our proposed data selection method worked well and consequently generated better quality data. In addition to that, our generated corpora are relatively smaller than their proposed methods. Our largest sub-corpora has 1M sentences, whereas B7 proposed three corpora which the smallest is three times larger than top6. Practically speaking, applying domain adaptation as well as using a large in-domain corpus in the work of~\citeasnoun{wang-etal-2017-sentence} caused translation performance to be only 1.5 BLEU points higher than top6. That would be negligible as we didn't benefit from any domain adaptation techniques.

\section{Discussions}
\label{sec:discussion}
In this section, first we discuss the statistical significance to evaluate the performance of MT models that trained on top1, top1+top2 and so on. Second, we want to investigate the training time of models trained using our proposed data and compare them with the training time of intended systems (S1, S2 and S3). Third, we show how different batch sizes affect the performance of training models in our research. Fourth, we want to investigate the impact of adding more data on the MT models' quality, more specifically, to see how much our mixing idea helped the models to improve translation quality.

\subsection{Statistical Significance}
\label{subsec:ss}

As mentioned earlier in Section \ref{sec:results_analayses}, we also computed pairwise statistical significance of scores shown in Table \ref{tab:indomain_data_selection} in terms of BLEU, TER and chrF2 scores by using bootstrap resampling and 95\% confidence interval for both test sets based on 1000 iterations and samples of 100, 200 and 300 sentences. According to experiment output, most results have a statistically significant difference except those system pairs listed in Table \ref{tab:ss_exception}. This evaluation reveals at some point models become similar. As such, at this point, the mixing sub-corpora (top6+top7, top7+top8, etc.) degrades the translation quality of MT systems. 

\begin{table}[ht]
\renewcommand{\arraystretch}{1.3}
\centering
\begin{adjustbox}{width=210pt,center}
\begin{tabular}{l|c|c|}
\cline{2-3} 
                                    & \textbf{Test Set 2010} & \textbf{Test Set 2011} \\ \hline
\multicolumn{1}{|r|}{\textbf{BLEU}}  & (Top4, Top5, 100)      & (Top5, Top6, 100)      \\ \hline
\multicolumn{1}{|r|}{\textbf{TER}} & \begin{tabular}[c]{@{}c@{}}(Top4, Top5, 100) \\ (Top4, Top5, 200)\\ (Top4, Top5, 300)\end{tabular} & - \\ \hline
\multicolumn{1}{|r|}{\textbf{CHRF2}} & -                    & -                  \\ \hline \hline
\end{tabular}
\end{adjustbox}
\caption{Results of system pairs that are \textbf{not} statistically significant (for $p < 0.05$). (TopX, TopY, N) means system TopX and TopY have \textbf{no} statistically significant difference based on N samples. When it comes to chrF2 and TER (Test Set 2011) we note that all results are statistically significant.  }
\label{tab:ss_exception}
%\vspace{-3mm}%Put here to reduce too much white space after your table
\end{table}

\subsection{Training Time}
\begin{table}[ht]
\renewcommand{\arraystretch}{1.3}
\centering
\begin{adjustbox}{width=310pt,center}
\begin{tabular}{l|c|l|c|l}
\hline
\hline
\multicolumn{1}{l|}{\textbf{Systems}} &
\textbf{\begin{tabular}[c]{@{}c@{}}Complete TT\\ D:H:M\end{tabular}} &
\multicolumn{1}{c|}{\textbf{Step}} &
\multicolumn{1}{l|}{\textbf{\begin{tabular}[c]{@{}c@{}}Best model TT\\ D:H:M\end{tabular}}}  &
\multicolumn{1}{l}{\textbf{Step}} \\ 
\hline
S1: ID            & 00:03:53   & 18,000 & 00:00:50   & 5,000   \\
S2: OOD           & 02:05:31   & 93,000 & 01:22:56   & 82,000  \\
S3: ID+OOD        & 01:13:30   & 39,000 & 00:10:00   & 29,000  \\ 
\hline
Top1              & 00:04:43   & 16,000 & 00:01:23   & 5,000   \\
Top2 + top1 + ... & 00:04:43   & 20,000 & 00:02:13   & 10,000  \\
Top3 + top2 + ... & 00:06:06   & 26,000 & 00:03:03   & 13,000  \\
Top4 + top3 + ... & 00:06:23   & 27,000 & 00:03:53   & 17,000  \\
Top5 + top4 + ... & 00:10:33   & 35,000 & 00:05:50   & 20,000  \\
Top6 + top5 + ... & 00:08:20   & 35,000 & 00:04:26   & 19,000  \\ 
\hline \hline
\end{tabular}
\end{adjustbox}
\caption{The training time (abbreviated as TT) of generated sub-corpora and the first category of compared systems.}
\label{tab:training_time}
\end{table}

As can be seen in Table \ref{tab:training_time}, a large baseline model (S2) with almost 31M sentences not only took 2 days, 5 hours and 31 minutes to be trained but also did not perform well in the context of in-domain translation – it obtained 25.8 and 30.7 BLEU score for test set 2010 and 2011, respectively. That means enormous training data is not always helpful for training in-domain MT systems. Moreover, according to Table \ref{tab:indomain_data_selection} data, the best performed mixed MT models are top5 and top6 for test sets 2011 and 2010, respectively, while, their training time is also considerably less than S2 and even S3. That resulted while S3 was mixed with parallel in-domain data, whereas proposed MT systems were trained without mixing with IWSLT in-domain corpus. The training time of top1 and other sub-corpora with no mixing procedure (Table \ref{tab:without_mixing}) is nearly the same, hovering around 1 hour and 20 minutes for finding the best model.

\subsection{Batch Size Effect}\label{sec:batch_size}
\begin{figure*}[htp]
	\centering 
	\includegraphics[scale=0.65]{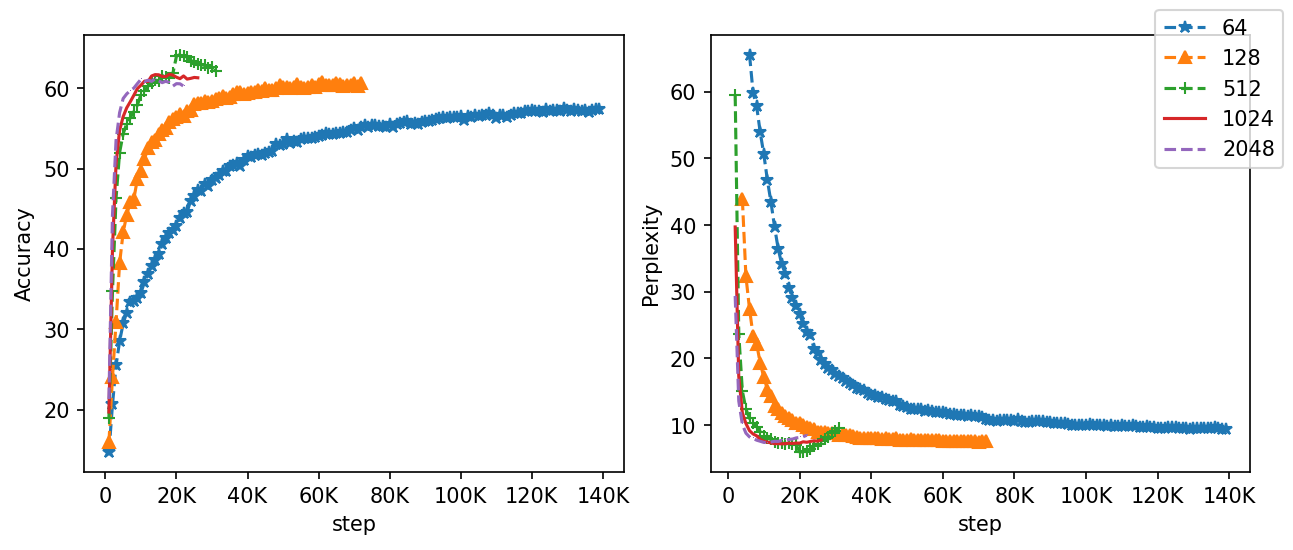} 
	\caption{The effect of different batch sizes on training the proposed NMT models. Left and right Figures show validation accuracy and validation perplexity per step, respectively.} 
	\label{fig:batch_sizes}
\end{figure*}

In this article, we use the hyperparameters defined by \citeasnoun{vaswani2017attention} to train our NMT models. However, the effects of different batch sizes on the models' performance were investigated as well in order to determine the most suitable one. This is because our work is pertinent to data selection. The size of selected data has a close correlation with batch sizes. As such, we need to choose a batch size that suits our models according to our sub-corpora size. To this end, we tested five different batch sizes including, 64, 128, 512, 1024 and 2048 for training on top1+..+top6 data. Figure \ref{fig:batch_sizes} depicts the accuracy and perplexity for the different batch sizes per step. 

These results show that the selected model with batch size 512 simultaneously reached the highest possible accuracy percentage (61.89\%) and the lowest perplexity score (7.08) within 21K steps. Following these experiments and the results shown in Figure~\ref{fig:batch_sizes} we decided to conduct all our experiments (see Section~\ref{sec:experiments}) with a batch size of 512. Further investigation is needed in order to define a correlation between (training) data size and batch size for optimal MT performance. We leave this for future work.

\subsection{Mixing Effect}
In our main experiments, we investigated the translation quality of systems trained on incremental data sets: top1 then top1+top2 and so on. In order to determine whether the quality achieved by the different systems is due to the quality of the data or the quantity (i.e. adding additional data), we trained 6 other models without combining (mixing) the data sets, i.e. a model trained only on top1, a model trained on top2 only and so on. Then we evaluated these models the same way as with the ones presented in Section~\ref{sec:experiments}. The results are shown in Table~\ref{tab:without_mixing}.
% \usepackage{multirow}
\begin{table}[ht]
\renewcommand{\arraystretch}{1.3}
\centering
\begin{adjustbox}{width=350pt,center}
\begin{tabular}{l|l|c|c|c|c|c|c}
\hline
\hline
\multirow{2}{*}{Systems} &
  \multirow{2}{*}{\begin{tabular}[c]{@{}l@{}}Sentence \\ Number\end{tabular}} &
  \multicolumn{3}{c|}{NMT- Test Set 2010} &
  \multicolumn{3}{c}{NMT- Test Set 2011} \\ \cline{3-8} 
 &
   &
  \multicolumn{1}{l|}{BLEU$\uparrow$} &
  \multicolumn{1}{l|}{TER$\downarrow$} &
  \multicolumn{1}{l|}{CHRF2$\uparrow$} &
  \multicolumn{1}{l|}{BLEU$\uparrow$} &
  \multicolumn{1}{l|}{TER$\downarrow$} &
  \multicolumn{1}{l}{CHRF2$\uparrow$} \\ \hline
Top1     & 179K     & 21.8      & 69.8       & 50.0           & 25.6       & 64.0       & 53.0  \\
Top2     & 179K     & 21.2      & 72.1       & 49.0           & 24.6       & 67.3       & 52.0  \\
Top3     & 179K     & 21.9      & 71.3       & 49.0           & 25.2       & 66.1       & 52.0  \\
Top4     & 179K     & 21.1      & 71.7       & 49.0           & 24.6       & 66.3       & 52.0  \\
Top5     & 179K     & 20.8      & 72.0       & 49.0           & 24.6       & 67.3       & 51.0  \\
Top6     & 179K     & 21.9      & 69.6       & 49.0           & 24.1       & 66.8       & 51.0  \\ 
\hline \hline
\end{tabular}
\end{adjustbox}
\caption{Results of in-domain data selection without mixing sub-corpora}
\label{tab:without_mixing}
%\vspace{-3mm}%Put here to reduce too much white space after your table
\end{table}

%Table \ref{tab:without_mixing} shows the performance of in-domain translation of NMT systems trained without mixing the selected sub-corpora. 
According to the three evaluation metrics, all systems performances are on par with system top1 except for minor differences that are unavoidable. That is, according to our experiment (shown in Figure~\ref{fig:closeness} in Section \ref{sec:data}) the centroids of selected sub-corpora (top1, top2, etc.) are very similar to centroids of in-domain test sets. This similarity between the centroid of top1 and the centroid of test sets 2010 and 2011 is even greater than others. To see how these numbers are interpreted, we computed pairwise statistical significance of BLEU score using bootstrap resampling and 95\% confidence interval~\cite{koehn-2004-statistical}. Table \ref{tab:SS_without_mixing} indicates whether the differences between two systems' BLEU scores are statistically significant (Y) or not (N). It shows that there was no statistically significant difference between results obtained with systems trained on top2, top3, ..., top6. However, the results obtained with top1 are statistically significant from all the rest. These are the sentences (from the OOD corpus) that are the most similar to the in-domain data. Table \ref{tab:SS_without_mixing} shows the statistical significance computed over both test sets and based on 1000 iterations and samples of 200 sentences. %Note that, Yes and No are abbreviated as 'Y' and 'N' in Table \ref{tab:SS_without_mixing}, respectively.

\begin{table}[ht]
\renewcommand{\arraystretch}{1.3}
\centering
\begin{adjustbox}{width=220pt,center}
\begin{tabular}{c|c|c|c|c|c|c}
\hline\hline
  & Top1 & Top2 & Top3 & Top4 & Top5 & Top6 \\\hline
Top1 &  & Y & Y & Y & Y & Y\\\hline
Top2 &  &  & N & N & N & N\\\hline
Top3 &  &  &  & N & N & N\\\hline
Top4 &  &  &  &  & N & N\\\hline
Top5 &  &  &  &  &  & N\\\hline
Top6 &  &  &  &  &  & \\\hline \hline
\end{tabular}
\end{adjustbox}
\caption{Results of statistically significant test for in-domain data selection without mixing sub-corpora (for $p < 0.05$).}
\label{tab:SS_without_mixing}
\end{table}

\begin{figure*}[htp]
  \centering
  \subfigure[]{\includegraphics[scale=0.45]{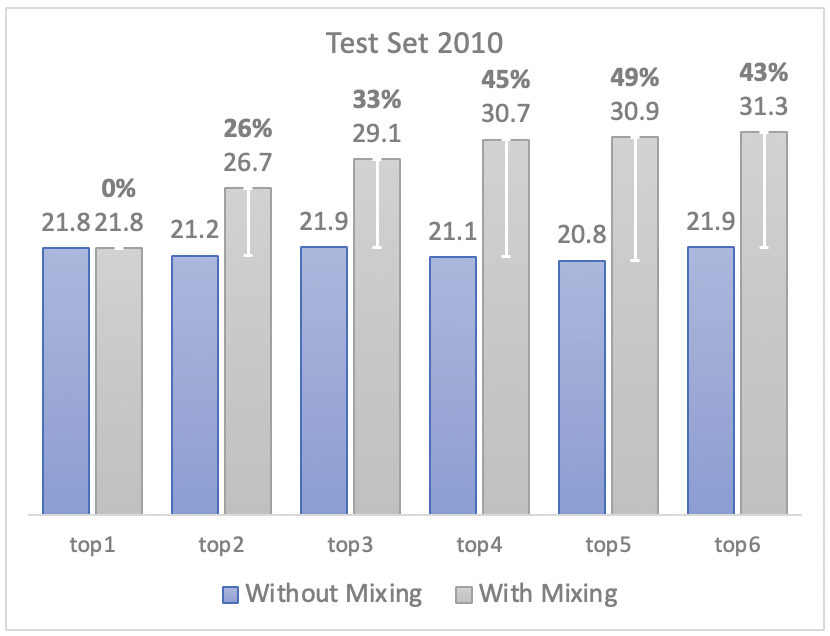}}
  \vspace{1.5mm}
  \subfigure[]{\includegraphics[scale=0.45]{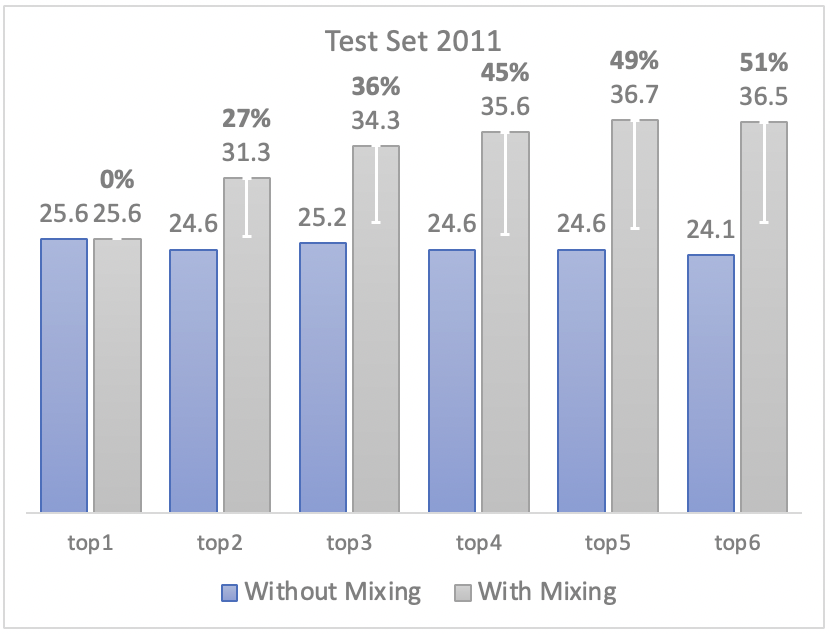}}
 \caption{Quality improvement (in percentage) after applying the mixing idea on NMT systems for test sets 2010 (a) and 2011 (b).}
 \label{fig:mixing_difference}
\end{figure*}

Figure \ref{fig:mixing_difference} also indicates the quality improvement (in percentage) of NMT systems that employed the mixture idea as compared to the original sub-corpora without being mixed. According to these figures, the mixing procedure enhanced the translation quality up to 49\% and 51\% for test sets 2010 and 2011, respectively. Overall, there is a gradual increase trend after mixing them with all preceding sub-corpora until the convergence point. That is, our objectives are (i) to mix as few sub-corpora as possible and (ii) at the same time increase the translation quality by avoiding the MT models being biased toward in-domain data. For example, top5 and top6 are convergence points for test sets 2011 and 2010, respectively. As such, we stop mixing sub-corpora once we reached our convergence points. For instance, the improvement rate for test set 2010 began with 26\%, then continued to 33\%, 45\% and eventually reached its peak by 49\% before dropping to 43\%. It is noteworthy that top1 had no improvement since we did not mix it with any other sub-corpora.

\section{Conclusion and Future Work}
\label{sec:conclusions}
In this paper, we presented a method to help the MT community to mitigate a lack of parallel in-domain corpora for many language pairs. Considering multiple standards such as generating high-quality data, designing a scalable architecture, having a reusable pipeline, we present a method for data selection, based on ranking, for the purposes of generating parallel in-domain data as well as domain adaptation. Given a large parallel corpus, the proposed method aims to select data that are semantically similar to a set of in-domain data. Typically, this method would be employed when there is little in-domain or only monolingual data, however, our method is generic and not restricted to the size of the in-domain data. The proposed selection pipeline is made of three main components: (i) a contextual sentence embedding component; (ii) a semantic search component and (iii) a ranking in-domain data component. 

We conducted experiments with different sizes of selected in-domain data. Our experimental results showed that selected parallel corpora generated through our data selection method can be applied directly for domain-specific NMT systems such that trained models outperformed the intended baselines and even their performances are comparable and at some points better than fine-tuned models. We note that our experiments used data up to top-6 as beyond top-6 a larger portion of the selected data becomes more dissimilar and would not contribute to the translation quality of the NMT systems. However, for other data and domains this threshold could be different and should be determined on a case-by-case basis. In our future work, we intend to employ our generated corpora in the context of domain adaptation by further training on a parallel in-domain corpus, which possibly boosts NMT systems' performance considerably. Furthermore, our proposed selection method shortens the training time and, at the same time, increases NMT translation quality compared to employing an out-of-domain corpus. We would also like to address other important research questions. First, we would like to compare the hard decision on $n$ -- the number of sentences that can be selected by our algorithm -- to a similarity threshold (which is not linked to any specific $n$). Another research direction would be to investigate the effects of our method on other language pairs and domains. 

The selected data and trained models are available at: \url{https://github.com/JoyeBright/DataSelection-NMT}.

%This implies that models trained on large data sets not only may not perform adequately for domain-specific MT systems, they also increase training time.

%\nocite{Sag} % items in your bibliography file that are not cited in your text

% add other references to the file bibliography.bib in this directory
\bibliographystyle{clin} 
\bibliography{bibliography}

\end{document}